\documentclass[10pt,letterpaper]{article}

\usepackage{cogsci}

\cogscifinalcopy 

\usepackage{color}
\usepackage{verbatim}
\usepackage{graphicx}
\usepackage{pslatex}
\usepackage{apacite}
\usepackage{float} 

\title{CogME: A Cognition-Inspired Multi-Dimensional Evaluation Metric\\ for Story Understanding}

\author{{\large \bf Minjung Shin\textsuperscript{1}(mjshin77@snu.ac.kr), Seongho Choi\textsuperscript{2}(shchoi@bi.snu.ac.kr),}\\
{\large \bf Yu-Jung Heo\textsuperscript{3}(yj.heo@kt.com), Minsu Lee\textsuperscript{4}(minsulee@snu.ac.kr)}\\
{\large \bf Byoung-Tak Zhang\textsuperscript{1,2,4}(btzhang@snu.ac.kr), Jeh-Kwang Ryu\textsuperscript{5}(ryujk@dongguk.edu)}\\
  \textsuperscript{1}Interdisciplinary Program in Cognitive Science, Seoul National University, \\
  \textsuperscript{2}Department of Computer Science and Engineering, Seoul National University,
  \textsuperscript{3}KT, \\
  \textsuperscript{4}AI Institute of Seoul National University(AIIS), 
  \textsuperscript{5}Department of Physical Education, Dongguk University\\
  \textsuperscript{1,2,4} Seoul, 08826, Republic of Korea, \textsuperscript{3} Seoul, 06763, Republic of Korea, \textsuperscript{5} Seoul, 04620, Republic of Korea}

\begin{document}

\maketitle

\begin{abstract}

We introduce CogME, a cognition-inspired, multi-dimensional evaluation metric for AI models focusing on story understanding. CogME is a framework grounded in human thinking strategies and story elements that involve story understanding. With a specific breakdown of the questions, this approach provides a nuanced assessment revealing not only AI models' particular strengths and weaknesses but also the characteristics of the benchmark dataset. Our case study with the DramaQA dataset demonstrates a refined analysis of the model and the benchmark dataset. It is imperative that metrics align closely with human cognitive processes by comprehending the tasks' nature. This approach provides insights beyond traditional overall scores and paves the way for more sophisticated AI development targeting higher cognitive functions. 

\textbf{Keywords:} 
artificial intelligence; video story understanding; video question answering; evaluation metric
\end{abstract}

\section{Introduction} \label{sec1}

In recent years, the development of artificial intelligence (AI) models, particularly pre-trained Large Language Models (LLMs) \cite{Vaswani2017_attention} and diffusion models \cite{Ho2020diffusion}, has made remarkable progress. 
These models have demonstrated impressive performance in creative tasks, including generating images \cite{rombach2022high}, videos \cite{singer2023makeavideo}, and narrative storytelling \cite{Brown2020GPT3}. 

However, skepticism remains about their ability to `understand,' as highlighted in recent discussions \cite{van2023ai, milliere2024philosophical}.
Indeed, while AI models often outperform humans in generating text and images, their performance in understanding does not reach their outstanding generative outputs \cite{west2023generative}.
This limitation becomes prominent in multi-modal AI models, where integrating and interpreting various data forms – such as image, video, and textual information – imposes considerable challenges.
In line with this limitation, video story understanding models demonstrate a significant gap compared to human story comprehension \cite{zhong2022-videoqa}.

Another key issue is the inadequacy of current evaluation metrics for AI models.
There are widespread arguments that existing metrics are too general and fail to provide a comprehensive analysis of these models. 
These metrics often rely on aggregate scores, which can obscure true model performance and hinder understanding the benchmark dataset's detailed features used for training \cite{Gundersen2018SOTA, burnell2023rethink}. 
This becomes increasingly pronounced in more complex tasks, like understanding video stories. Consequently, there is an urgent need for a new method that aligns with human cognition for a more fine-grained assessment of AI model performance. 
Addressing this pressure would require a thorough understanding of the intrinsic nature of given tasks and the evaluation target.

\begin{figure}[t]
\begin{center}
\includegraphics[width=\columnwidth]{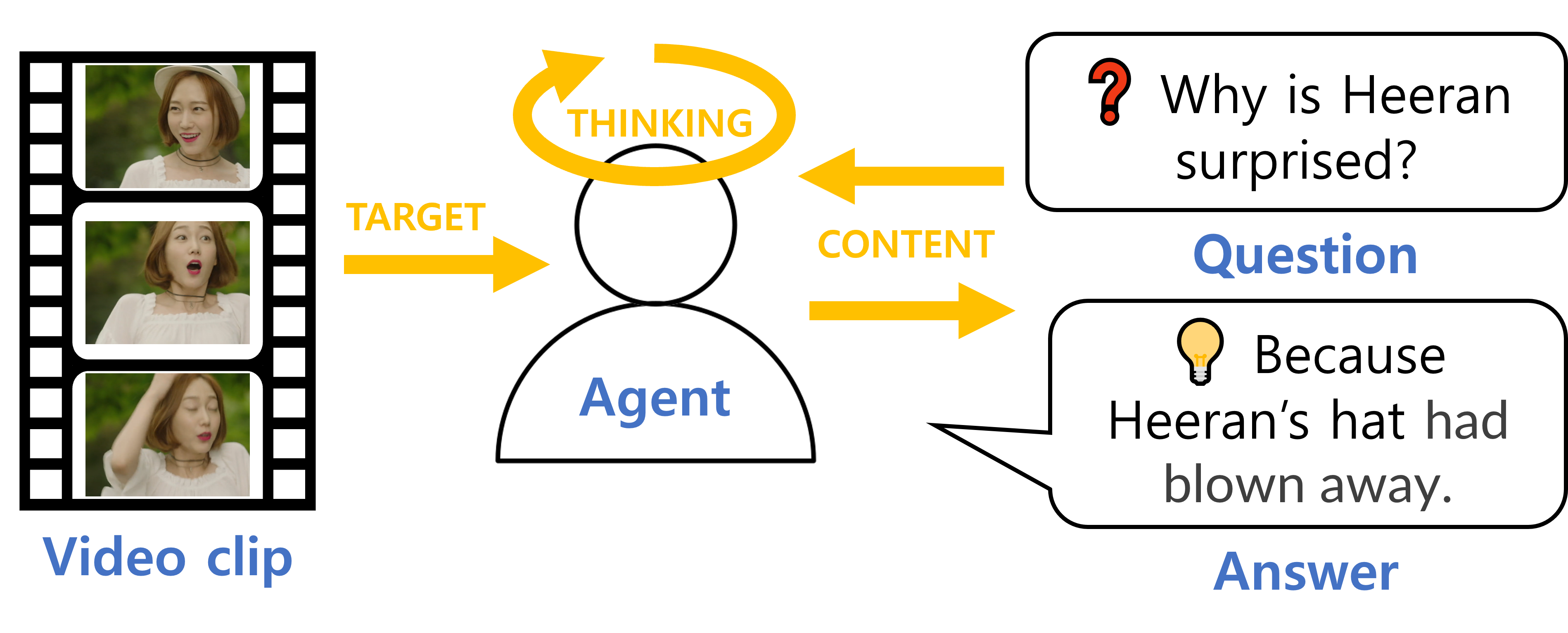}
\end{center}
\caption {An illustration of CogME framework for an example of DramaQA dataset. It shows a situation in which an \textit{Agent} predicts the \textit{Answer} from the given \textit{Video clip} and \textit{Question}. Orange arrows indicate the process involves three story understanding components: TARGET, CONTENT, and THINKING. }
\label{fig:1}
\end{figure}

In light of this need, we have proposed a novel evaluation metric for the story understanding model, named \textbf{Cog}nition-inspired \textbf{M}ulti-dimensional \textbf{E}valuation \textbf{(CogME)}. CogME is designed to evaluate through a unique lens: A specific breakdown of the questions. The breakdown is grounded in multi-dimensional criteria that consider human thinking strategy and story elements.

The unique design is based on the following proposition: \textit{If an agent answered a specific question appropriately, it means that ``The agent understood the CONTENT of the TARGET through a way of THINKING." It also means ``The question required the agent's THINKING about the CONTENT of the TARGET."} 
Our approach analyzes the context intricately to provide a richer explanation of how AI models tackle each query. 
This framework not only identifies the strengths and weaknesses of AI in handling various questions but also highlights the dataset's distinctive features.

We also present the results from an in-depth case study using DramaQA, a representative benchmark for video story understanding, along with its baseline model.
Fig.~\ref{fig:1} shows how CogME is applied to DramaQA. 
Our findings confirm that CogME enables a thorough and systematic evaluation of both benchmark datasets and AI models.

\section{Related Work}\label{sec2}
\subsection{Narrative Comprehension of Human} \label{subsec21}
\subsubsection {Cognitive science studies on narrative comprehension}, conducted through various reading comprehension tasks, have yielded seminal findings. It was found that people tend to prioritize the main aspects of a story over individual parts, indicating a higher level of cognitive engagement with narrative comprehension \cite{THORNDYKE197777}. 
The concept of \textit{schema} \cite{bartlett1933remembering, brewer1985story} in memory reveals that recall is an active process influenced by personal and cultural contexts, emphasizing the critical roles of context and prior knowledge in comprehension \cite{bransford1972contextual, Graesser1994ConstructingID}. Mental representations such as \textit{scripts} and \textit{schemata} were denoted as crucial cognitive structures aiding comprehension and inference-making \cite{schank1975, rumelhart1980schemata}.

From these research bases, the \textit{Situation model} and \textit{Event-indexing model} laid the groundwork for analyzing human story comprehension, providing a structured basis for evaluation. 
The situation model constructs detailed mental representations encompassing events, characters, and settings \cite{zwaan1998situation}. 
The event-indexing model emphasizes five independent dimensions of understanding when reading: time, space, character, causality, and motivation in the narrative context \cite{zwaan1995}.

\subsubsection {Research on understanding video stories} is relatively scarce compared to reading comprehension despite the rapidly increasing importance of video narratives in our daily lives. 
Since reading and viewing are distinct tasks with different cognitive loads \cite{jajdelska2019picture, cohn2020your}, it is undesirable to replicate reading comprehension structures for video comprehension \cite{gibson1979ecological, hochberg1996perception}.

Recent research focuses on understanding video narratives in real-world contexts. 
Specifically, top-down approaches involve observing brain activity, not in controlled or fragmented experimental videos, but rather in watching typical movie scenes \cite{baldassano2018representation, song2021cognitive}. 
They contrast with conventional cognitive psychology studies, which primarily employ a bottom-up approach focusing on the segmented visual and audio stimuli \cite{tan2018psychology}. 
These integrated approaches emphasize considering both perceptual and narrative aspects to understand how people interpret video narratives in real-world situations.

\subsection{Evaluation of Machine Comprehension}\label{subsec22}

\subsubsection {Machine Reading Comprehension (MRC)} is the most prominent within machine comprehension \cite{hirschman1999deep}. 
Question-answering (QA) has been widely adopted to evaluate text understanding of MRC models. \cite{burges2013towards, baradaran2022survey}.
However, efforts to access the MRC models systematically have only recently been made.
Dunietz et al. argued that existing MRC metrics lack clarity and could be improved by using templates derived from the definition of comprehension \cite{dunietz-etal-2020-test}.
To address this issue, they employed the \textit{Event-indexing model} from human studies \cite{zwaan1995} to posit four elements that machines should incorporate for better reading comprehension: place, time, causality, and motivation.
Similarly, Weston et al. demonstrated empirical test results of multiple AI agents’ textual understanding abilities with structured QA skillsets categorized twenty types of questions for understanding and reasoning with text \cite{weston2015toyquestion}. 
However, these tests are limited to being constructed with fragmentary and artificial descriptions. 
Further research is needed before their method can be applied at the level of complexity and richness typically found in everyday human storytelling.

\subsubsection {Video understanding models} have yet to keep pace with the growing demand, even though video-based storytelling has recently emerged as one of the most prominent forms of media content. Developing AI that can understand video stories is challenging, given that it requires an all-inclusive process to analyze images, scripts, and sounds with temporal dependencies, natural language, and various levels of reasoning \cite{bebensee2021}. 
Despite the difficulty, several efforts to develop video understanding AI have centered on large-scale video datasets \cite{Tapaswi2016MovieQA, lei2018tvqa, Yu2019ActivityNetQA, garcia2020knowit, Choi2021dramaqa, Yang_2021_JustAsk}, but the technologies do not extend beyond simple image processing tasks such as detecting or tracking objects.

From the evaluation perspective, the most prevalent approach involves building massive QA datasets, leveraging open-ended or multiple-choice QAs for AI training and testing \cite{PatelPS20}. 
However, existing evaluation methods heavily count on unidimensional metrics \cite{aafaq2019video}, such as basic QA accuracy scores for the multiple-choice QA datasets, which often fall short in providing a comprehensive explanation of the model's performance.
In the case of open-ended QA, the automatic evaluation primarily depends on n-gram-based sentence similarity measures such as BLEU \cite{papineni2002bleu}, METEOR \cite{banerjee2005meteor}, ROUGE \cite{lin2004rouge}, and CIDEr \cite{vedantam2015cider}. 
These metrics, focusing on the similarity between the reference and generated sentences, frequently do not align with human judgments, reflecting a significant disconnect.
These automatic metrics are usually insufficient to provide a refined interpretation like the strengths and weaknesses of the AI model \cite{nema2018towards}.

On the other hand, there are several efforts in human evaluations to appraise the effectiveness of automatic metrics in accurately assessing AI performance. \cite{chen2019evaluating, garcia2020knowit}. 
While these efforts indicate that human evaluations are beneficial for assessing AI performance, they also reveal a significant challenge: the fine-grained evaluation is not effectively achieved in proportion to the resources and costs involved in the evaluation process.

To sum up, evaluating AI solely based on its QA accuracy or similarity to human performance is inadequate, especially for tasks with high complexity, such as video story understanding. Accordingly, a structured framework that meticulously analyzes both the nature of the understanding process and the unique features of the medium is needed. In response to this, we propose a new metric that not only reflects the existing frameworks of human story comprehension but also incorporates the distinct attributes of video storytelling.

\begin{table}[t]
\centering
\caption{The sub-components within TARGET}
{\renewcommand{\arraystretch}{1.1}%
\begin{tabular}{p{0.24\columnwidth} | p{0.65\columnwidth}}
\hline\hline
    \textbf{Elements} & \textbf{Definition} \\
\hline\hline
    \texttt{Character}
    & Information of individuals featured in the video. \\
    \hline
    \texttt{Object} 
    & Items and body parts featured in the video.\\
    \hline
    \texttt{Place}	
    & Spatial information of the story in the video.\\
    \hline
    \texttt{Conversation}	
    & Characters' dialogues, monologues, speech sounds, and text messages.\\	
    \hline
    \texttt{Behavior}
    & Movements and actions of the subjects in the video. \\
    \hline
    \texttt{Event}	
    & Information about what happened in the video. \\
    \hline
    \texttt{Emotion}	
    & The feelings expressed by the subject in the video.\\
    \hline
    \texttt{Commonsense}	
    & Concepts and knowledge that people universally accept in a given culture.\\
\hline\hline
\end{tabular}
\label{table:1}}
\end{table}

\begin{table}[t]
\centering
\caption{The sub-components within CONTENT}
{\renewcommand{\arraystretch}{1.1}%
\begin{tabular}{p{0.24\columnwidth} | p{0.67\columnwidth}}
\hline\hline
    \textbf{Elements} & \textbf{Definition} \\
\hline\hline
    \texttt{Identity}	
    & Personal information of subjects or names of objects in the story. \\
    \hline
    \texttt{Feature}	
    & Characteristics, traits, or atmosphere of subjects and/or objects. \\
    \hline
    \texttt{Relationship}	
    & The relationships between two or more targets. \\
    \hline
    \texttt{Means}	
    & Instruments or methods used to achieve a particular purpose. \\
    \hline
    \texttt{Context}	
    & Story-line revealed through the conversations or interactions between characters. \\
    \hline
    \texttt{Sequence}	
    & Related events with time series and the changes before and after. \\
    \hline
    \texttt{Causality}	
    & Causes and consequences of a particular change: natural or mechanical. \\
    \hline
    \texttt{Motivation}	
    & Changes resulting from actions involving personal preferences or intentions. \\
\hline\hline
\end{tabular}
\label{table:2}}
\end{table}

\begin{table}[t!]
\centering
\caption{The sub-components within THINKING}
{\renewcommand{\arraystretch}{1}%
\begin{tabular}{p{0.2\columnwidth} | p{0.71\columnwidth}}
\hline\hline
    \textbf{Elements} & \textbf{Definition} \\
\hline\hline
    \texttt{Recall}	
    & Retrieving or recollecting factual information in the scene.\\
    \hline
    \texttt{Grasping}	
    & Perceptions or interpretations of the scene with temporal and spatial changes.\\
    \hline
    \texttt{Reasoning}	
    & Making logical judgments from circumstantial evidence not direct observations.\\
\hline\hline
\end{tabular}
\label{table:3}}
\end{table}

\section{New Evaluation Paradigm Based on the Understanding Processes of Humans}\label{sec3}

To evaluate understanding competence thoroughly, we developed multifaceted criteria integrating video narrative elements and thinking strategies involving queries. 
In analyzing the story elements provided by the video, we have adapted the Situation Model \cite{zwaan1998situation} and the Event-Indexing Model \cite{zwaan1995} and expanded them to better suit video narratives.
Regarding human thinking strategies, we referred to Bloom's Taxonomy, widely accepted as a representative framework demonstrating the hierarchical structure of cognitive processes \cite{bloom1956taxonomy, anderson2001taxonomy}.
Each level of Bloom's taxonomy represents a different cognitive skill, ranging from the basic recall of facts and grasping details to reasoning hidden matter and, ultimately, evaluation and creation.

\begin{figure*}
\begin{center}
\includegraphics[width=\textwidth]{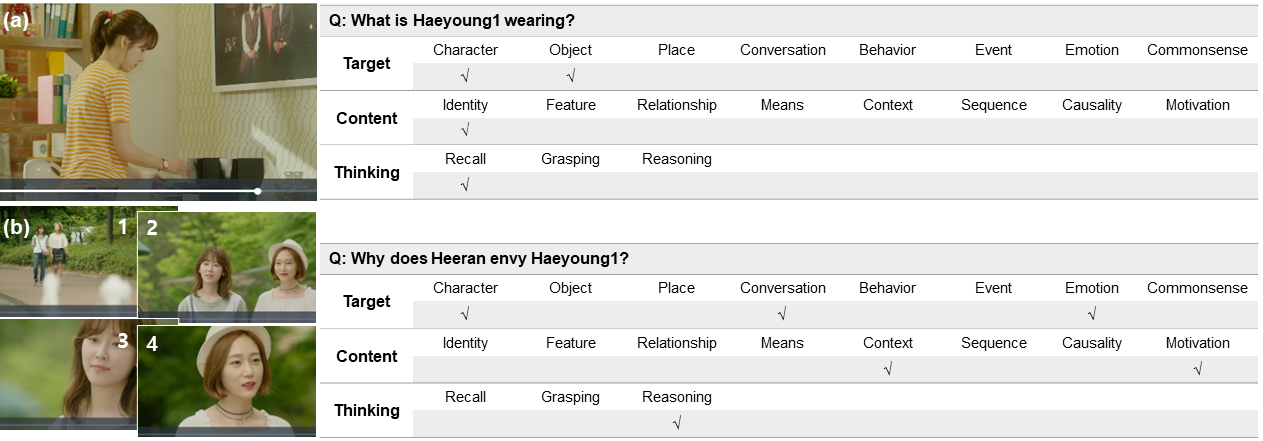}
\end{center}
\caption{Examples of tags applied to questions \textbf{(a)} Cases of tagged to questions in \textit{shot}-level video, which require simple recall. \textbf{(b)} Cases of tagged to questions in \textit{scene}-level video, which require comprehensive reasoning.} 
\label{fig:annotation}
\end{figure*}

Drawing on insights from previous models in cognitive science, we have developed a multi-dimensional metric, CogME, which consists of three key components: TARGET, CONTENT, and THINKING. 
\textbf{TARGET} refers to the information perceived by watching the video, \textbf{CONTENT} to the knowledge acquired through the target information, and \textbf{THINKING} to the cognitive process of deriving knowledge from the information.
The three components, representing cognitive processes, integrate to interpret the story elements presented in the video, succinctly expressed as \textit{``I understand the CONTENT of the TARGET through a way of THINKING.”}

The sub-components of TARGET and CONTENT were determined by analyzing the story elements necessary for understanding video stories.
The sub-components of THINKING were decided to align with the range required among human thinking strategies.
Tables~\ref{table:1} --~\ref{table:3} demonstrate the sub-components of TARGET, CONTENT, and THINKING.

\section{Materials and Methods}\label{sec4}
\subsection{Application to VideoQA Dataset: DramaQA}\label{subsec41}
This study evaluated the DramaQA dataset, which included $\sim$16k human-generated QA pairs closely centered around the narrative of a TV series \textit{Another Miss Oh,} along with character-level annotations \cite{Choi2021dramaqa,bebensee2021}. 
The character-centered annotations and five-option multiple-choice QA pairs in DramaQA were generated by approximately two to five trained human annotators using a consistent manual for all 18 episodes.
The dataset was designed to reflect various narrative elements in the questioning stage, from seeking simple information from the video to reasoning complex causality about the stories \cite{heo2019constructing}. 
The baseline model of DramaQA, i.e., the Multi-level Context Matching (MCM) model \cite{Choi2021dramaqa}, was trained with the 1st to 12th episodes, validated with the 13th to 15th episodes and tested with the 16th to 18th episodes.

\subsection{Annotating the Understanding Components} \label{subsec42}

To determine what information, knowledge, and thinking strategies are required for answering questions, we annotated 4,385 questions from episodes 13th to 15th using the understanding sub-components defined in the CogME framework.
Two specialists in cognitive science elaborately analyzed the given videos and questions, tagging the required sub-components in each question. 
To ensure consistent annotation tagging, both individuals followed the same predefined manual and resolved any discrepancies in inter-rater annotations through discussion to reach a consensus.

Fig.~\ref{fig:annotation} illustrated two examples of this annotation.\footnote{The DramaQA dataset features memory capacity criteria related to the length of the video segments. The criteria include two types of video clip: 1) \textit{shot}-level video without camera cut, spanning a few seconds, and 2) \textit{scene}-level video with multiple events in a single location, spanning a few minutes}
Fig.~\ref{fig:annotation}(a) shows an example of a question answerable by simply recalling a single cue from a \textit{shot}-level video. It involves identifying clothing-related information about the only person in the shot. 
In contrast, Fig.~\ref{fig:annotation}(b) illustrates a complex scenario where tagging sub-components in a \textit{scene}-level video highlights a different level of narrative understanding from the example shown in Fig. \ref{fig:annotation}(a). 
During a three-minute and six-second runtime, two characters are shown walking in the park and chatting. 
To answer the question, the agent must not only recall and grasp the content of the conversation but also infer why Heeran said to envy Haeyoung, which is not directly mentioned in the dialog.

For THINKING strategies, we assumed that higher cognition encompasses lower ones based on the hierarchical Bloom's taxonomy \cite{bloom1956taxonomy}, so we labeled only the highest thinking component.
Regarding the TARGET and CONTENT of the question, we tagged up to three if multiple sub-elements were involved in a single question. 

\begin{figure*}[t]
    \centering
    \includegraphics[width=1.0\textwidth]{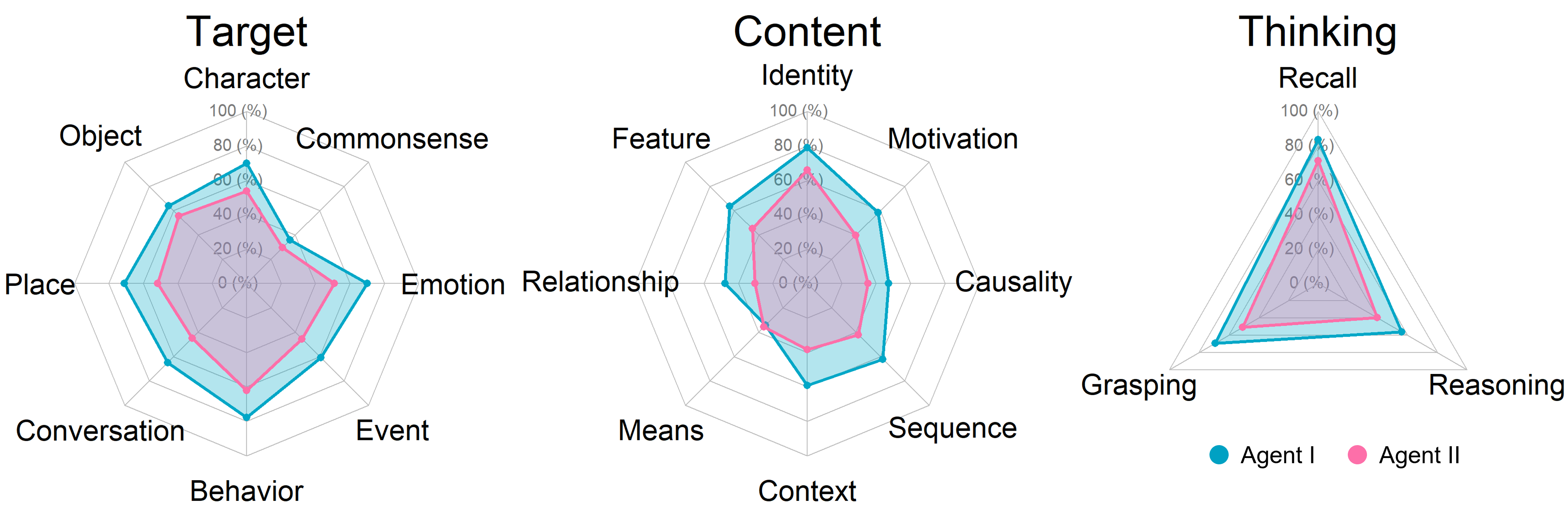}
    \caption{Performance profiles of two models. The vertex of each polygon represents the ratio (\%) of correct predictions for the DramaQA dataset. Each radar plot represents TARGET\textbf{(left)}, CONTENT\textbf{(middle)}, THINKING\textbf{(right)} component. Light blue areas indicate the performance profiles of Agent I (MCM model), and pink areas display the performance profiles of Agent II (MemN2N model).}
    \label{fig:cogme2}
\end{figure*}

\subsection{Scoring the Questions and Prediction Results} \label{subsec43}

In the context of the logical complexity of the THINKING module, a weight of \*2 and \*3 was assigned to \texttt{grasping} and \texttt{reasoning}, respectively, assuming grasping includes the simple recall and the reasoning process retains the recall and grasping. 
All tagged sub-components were multiplied by the weight given to the THINKING component, as the depth of the thinking strategy determines the overall difficulty of the question.
In the example shown in Fig.~\ref{fig:annotation}, for question in Fig.~\ref{fig:annotation}(a), which requires \texttt{recall}, all labeled elements are worth 1 point, while for question in Fig.~\ref{fig:annotation}(b), which requires \texttt{reasoning}, all labeled elements are worth 3 points. 

Even identical questions can be endowed with different weights. This variance depends on the diverse contexts of the given video, ultimately affecting the complexity of the question. 
Considering the question, `Who is smiling?' if there is only one character in the video, the prediction can be made simply by recalling who it is. 
However, when there are two or more characters, the prediction requires a higher level of thinking, which involves discriminating the smiling character from the others and then identifying that character. 
This grasping process certainly includes recalling relevant details.

The model's accuracy for each sub-component was calculated by scoring correct predictions as 1 and incorrect ones as 0. 
We then computed the overall success rate by comparing the number of correct predictions to the total attempts. 
To address the imbalanced frequency of each sub-component, the achievement rate was expressed as a percentage.

The analysis was conducted in R 4.1.1 \cite{r_manual} environment for arithmetic computing and visualization. The polygonal profiles were produced using the fmsb 0.7.1 package \cite{r_fmsb_manual}.

\subsection{Comparison of Multiple Agent Performances} \label{subsec44}
In additional experiments, two agents trained with the same dataset were evaluated with CogME to compare their performances. The two models were the MCM model \cite{Choi2021dramaqa} (Agent I) and the baseline model of MemN2N \cite{Sukhbaatar2015N2N} (Agent II)\footnote{It should be noted that these experiments were not meant to identify the model with the better performance but only to compare them objectively. Accordingly, in the \textbf{Results} section, we refer to the models as Agent I and Agent II instead of their names}.
The two models were examined after being trained on the same dataset, DramaQA. Their performance profiles were generated using CogME.      

\section {Results} \label{sec5}

\subsection{Evaluating Model Performances} \label{subsec51}

For a fine-grained analysis of model performance, the model's accuracy for each sub-component defined by CogME was scored based on correct predictions, and each accuracy was calculated by comparing correct predictions to the total attempts.

Fig.~\ref{fig:cogme2} shows the multi-dimensional accuracy profile for each understanding component obtained from applying CogME to the dataset. 
The profile indicates that the Agents demonstrate varying levels of competence across different sub-components, as depicted by the uneven shape of the polygons.
For example, based on Agent I, \texttt{Identity} of CONTENT shows an accuracy of 79.1\%, while \texttt{Means} only achieves 34.3\% accuracy.

Additionally, we observed distinct disparities when comparing the results of two different models, Agent I (MCM model, represented by a light blue polygon) and Agent II (MemN2N model, represented by a pink polygon).
When the two models were trained on the same dataset, the overall correct prediction rates were 73.4\% for Agent I and 58.7\% for Agent II, indicating a difference of 14.7\%. 
This difference clearly illustrates that, as is often the case in many benchmark analyses, the MCM model specialized for the DramaQA dataset outperforms the MemN2N model, which primarily targets natural language processing.

However, in a detailed breakdown according to CogME's criteria, this discrepancy is not uniform across all areas but varies by specific factors, as shown in Fig.~\ref{fig:cogme2}.
For instance, Agent II leads by approximately 1\% in questions requiring the identification of the \texttt{Means}, while the gap extends to 20\% in questions involving the \texttt{Conversation}. 
This discrepancy underscores the significance of the CogME metric in providing an in-depth understanding of each model's performance, a potential that has yet to be fully explored in this context.

\subsection{Analyzing Questions in DramaQA Dataset} \label{subsec53}

\begin{figure}[t]
\centering
\includegraphics[width=1.0\columnwidth]{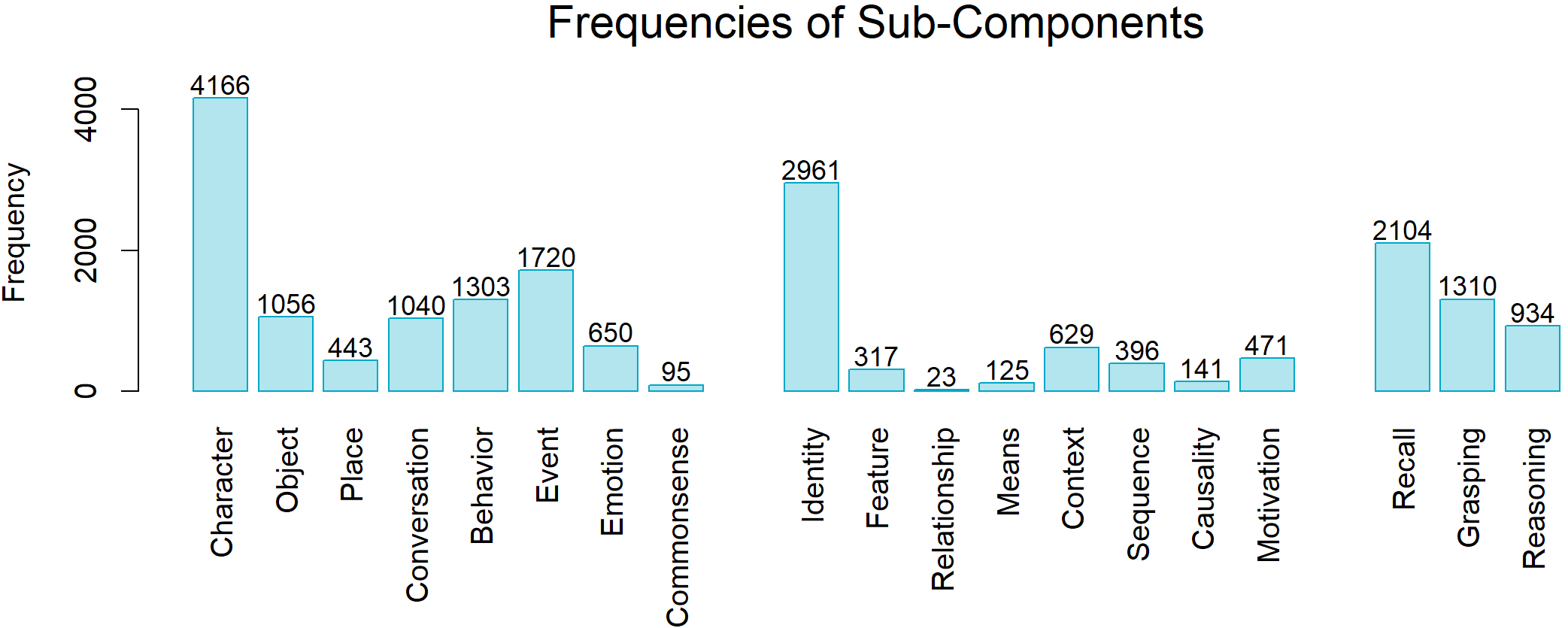}
\caption{Frequencies of sub-components tagged in the questions of the DramaQA dataset. Each bar shows the number of times a sub-component was labeled out of 4,385 questions.}
\label{fig:freq}
\end{figure}

Alongside performance profiling, annotations based on CogME enable us to figure out the dataset's features in terms of data distribution. 
Fig.~\ref{fig:freq} shows the distribution of sub-components tagged for the questions in the DramaQA dataset, which characterizes the benchmark.

The uneven distribution of sub-components reflects an unequal consideration of aspects of story comprehension during the dataset creation phase.
The prominent bars indicate that the dataset is heavily skewed towards questions that ask superficial information, like \textit{recalling a character's identity}.

Moreover, the sub-components significantly underrepresented in the dataset tend to align with lower accuracy in model performance.
Notably, all four elements that appeared less than 5\% in the dataset (i.e., \texttt{Commonsense, Relationship, Means} and \texttt{Causality}) showed low accuracies that are below 50\%: \texttt{Commonsense} (35.8\%), \texttt{Relationship} (47.8\%), \texttt{Means} (34.3\%), \texttt{Causality} (47.2\%).

\section {Discussion} \label{sec6}

In this study, we introduced a novel framework, CogME, centered around the features of the posed questions.
This approach is grounded in structured metrics that consider human thinking strategies and story elements using a top-down perspective. 
Unlike conventional AI evaluation methods, which emphasize overall scores that lead to a lack of robust evaluation, CogME provides a multi-dimensional quantified profile for AI models.
This profile provides insight into the model's strengths and weaknesses in understanding abilities by applying the metric to an existing dataset.

Our comparison of the two models using the CogME metric revealed detailed dissimilarities that their aggregate scores could not explain. 
These variations highlight the importance of a multi-dimensional evaluation approach for accurately assessing AI models' capabilities. 
This assessment is expected to apply to both machines and humans, providing a comprehensive quantification of the agents' levels of understanding. \cite{lee2023VTT}.

Furthermore, CogME's fine-grained evaluation not only assesses the AI models but also offers an analysis of the benchmark dataset, providing deeper insights into the models' capabilities. 
For instance, the observed link between low frequency and low accuracy \footnote{Although we only analyzed questions from the validation set, we assumed that the CogME profile of the training set would be similar based on the preliminary analysis that demonstrated a similar distribution of question types across the datasets. \cite{Choi2021dramaqa}}, indicates that learning deficiencies can impact QA performance, highlighting the need for a more balanced dataset covering various aspects of narrative comprehension.
Moreover, as noted in the \textbf{Narrative Comprehension of Human} section, people focus on a story's central aspects rather than individual instances \cite{THORNDYKE197777}.
However, our analysis reveals that this dataset predominantly collected fragmentary information instead of emphasizing the story's central or structural elements.

This insight leads to the establishment of a CogME framework as a guideline for designing new datasets.
Generating a massive QA dataset makes it challenging to ensure sufficient variety in question types and sub-components, as seen in Fig.~\ref{fig:freq}.
These maldistribution issues have been noted not only in DramaQA but also in many other QA datasets \cite{Garcia-Molina2016crowd}.
In this context, the CogME framework could serve as a theoretical foundation for proper data allocation in datasets, whether through crowdsourcing or automatic question generation. 

We acknowledge that a challenge in our study is that sub-components were tagged manually in the provided questions and videos.
It was inevitable to capture the elements comprehensively, as even the identical questions can vary in different video contexts (see \textbf{Scoring the Questions and Prediction Results} section). 
Despite being cumbersome, manual annotation ensures accurate evaluation by aligning with the nuanced content of the videos and related queries.
However, in the future, using a multi-modal classification model to automatically annotate sub-components in CogME could streamline the evaluation process. 
Such automation would not only simplify the evaluation process but also allow for the scalability of larger and more complex datasets.

Additionally, scoring multiple-choice questions can result in some information loss. 
According to our annotating, even if the model correctly recognized the \texttt{Character}'s \texttt{Identity} but failed to infer other information, like \texttt{Emotion}, it would score zero for that question, including the \texttt{Character}'s. 
By incorporating rubric-like methods used in pedagogy \cite{Brookhart2018rubric}, we argue that this metric could be adapted to other tasks like open-ended or fill-in-the-blank tests, summaries, and rewriting, which can be analyzed through understanding sub-components \cite{lee2023VTT}.

In conclusion, this study introduces the CogME framework, offering a multi-dimensional analysis focusing on story understanding that surpasses the limitations of overall scores, like accuracy rates. CogME's potential extends to various AI tasks and dataset designs, suggesting its adaptability and utility in advancing AI assessment toward more nuanced and sophisticated dimensions. This work also establishes a new benchmark, paving the way for more comprehensive approaches to developing AI agents.

\section{Acknowledgments}
We deeply thank the reviewers for providing kind and helpful comments. This work was partly supported by Institute of Information \& communications Technology Planning \& Evaluation (IITP) grants (No. 2017-0-01772, VTT | No. 20220-00951, LBA) and by the National Research Foundation of Korea (NRF) grant (No. RS-2024-00358416,  AutoRL) funded by the Korea Government (MSIT).

\bibliographystyle{apacite}

\setlength{\bibleftmargin}{.125in}
\setlength{\bibindent}{-\bibleftmargin}

\bibliography{cogsci_CogME}

\end{document}